\documentclass[journal]{IEEEtran}
\ifCLASSINFOpdf
\else
\fi
\usepackage[utf8]{inputenc} 
\usepackage[T1]{fontenc}    
\usepackage{hyperref}       
\usepackage{url}            
\usepackage{booktabs}       
\usepackage{amsfonts}       
\usepackage{nicefrac}       
\usepackage{microtype}      
\usepackage{graphicx}
\usepackage{amsmath}
\usepackage{amsthm}
\usepackage{wrapfig}
\usepackage{bm}
\usepackage{xcolor}
\usepackage{floatrow}
\usepackage{caption}
\usepackage{enumitem}
\usepackage{multirow}
\newfloatcommand{capbtabbox}{table}[][\FBwidth]

\newcommand{\algname}{\textsc{RegNet }}
\newcommand{\algnamens}{\textsc{RegNet}}
\newcommand{\auxname}{audio forwarding regularizer }
\newcommand{\auxnamens}{audio forwarding regularizer}

\newcommand{\e}[1]{{\small $#1$}}

\def\etal{{\em et al.\/}\,}

\def\mr{\textcolor{black}}
\def\aq{\textcolor{black}}

\begin{document}
	%
	\title{Generating Visually Aligned Sound from Videos}
	%
	%
	%
	
	\author{
		Peihao Chen,
		Yang Zhang,
		Mingkui Tan,
		Hongdong Xiao,
		Deng Huang, and
		Chuang Gan
		\thanks{
This work was partially supported by 
Science and Technology Program of Guangzhou, China under Grants 202007030007,
National Natural Science Foundation of China (NSFC) 61836003 (key project),
Program for Guangdong Introducing Innovative and Enterpreneurial Teams 2017ZT07X183, 
Fundamental Research Funds for the Central Universities D2191240.	
		}
		\thanks{
			P. Chen is with the School of Software Engineering, South China University of Technology, Guangzhou 510640, China, and also with the Pazhou Laboratory, Guangzhou, 510335, China
			(e-mail: phchencs@gmail.com).}
		\thanks{
			 M. Tan, H. Xiao, and D. Huang are with the School of Software Engineering, South China University of Technology, Guangzhou 510641, China (e-mail: mingkuitan@scut.edu.cn; xiaohongdonghd@gmail.com; im.huangdeng@gmail.com).
		}
		\thanks{
			Y. Zhang and C. Gan are with the MIT-IBM Watson AI Lab, Cambridge, MA 02142, USA (e-mail: yang.zhang2@ibm.com; ganchuang1990@gmail.com).
		}
	
		\thanks{P. Chen and Y. Zhang contributed equally to this paper.}
		\thanks{ M. Tan is the corresponding author.}
	}
	
	%
	%

	\markboth{Journal of \LaTeX\ Class Files,~Vol.~14, No.~8, July~2020}%
	{Shell \MakeLowercase{\textit{et al.}}: Bare Demo of IEEEtran.cls for IEEE Journals}
	%



	\maketitle
	
	\begin{abstract}
		We focus on the task of generating sound from natural videos, and the sound should be both temporally and content-wise aligned with visual signals. 
		This task is extremely challenging because some sounds generated \emph{outside} a camera can not be inferred from video content. The model may be forced to learn an incorrect mapping between visual content and these irrelevant sounds.
		To address this challenge, we propose a framework named \algnamens. In this framework, we first extract appearance and motion features from video frames to better distinguish the object that emits sound from complex background information. We then introduce an innovative \auxname that directly considers the real sound as input and outputs bottlenecked sound features. Using both visual and bottlenecked sound features for sound prediction during training provides stronger supervision for the sound prediction. The \auxname can control the irrelevant sound component and thus prevent the model from learning an incorrect mapping between video frames and sound emitted by the object that is out of the screen.
		During testing, the \auxname is removed to ensure that \algname can produce purely aligned sound only from visual features.  Extensive evaluations based on Amazon Mechanical Turk demonstrate that our method significantly improves both temporal and content-wise alignment.  Remarkably, our generated sound can fool the human with a 68.12\% success rate. Code and pre-trained models are publicly available at \href{https://github.com/PeihaoChen/regnet}{https://github.com/PeihaoChen/regnet}.
		
	\end{abstract}
	

	%
	\IEEEpeerreviewmaketitle

	\section{Introduction}
	%
	%
	%
	%
	
	
	
	\IEEEPARstart{V}{arious} visual events in our daily life are usually accompanied by different sounds.  Because visual events and sounds are correlated, a person can instinctively infer sounds by observing visual events.  In this paper, we will address the task of deriving sound from silent videos, which is beneficial to many real-world applications, such as video editing automation, generating sound for silent film, and assistance for people with visual impairment.
	
	The \emph{alignment} to the corresponding video is an essential characteristic for plausible generated sound.
	Specifically, a visually aligned sound should have two aspects.  The first aspect is the \emph{temporal alignment}, \emph{i.e.}, synchronization.  A temporally-aligned approach should produce a sound event at exactly the time or soon after the time when the corresponding visual event occurs.  For example, a dog barking should be synchronized with the visual event in which the dog opens its mouth and leans forward.
	The second aspect of the alignment is the \emph{content-wise alignment}. A sound that is content-wise aligned should produce only the types of sound that correspond to what is happening in the video.  For example, when a dog opens its mouth, we would expect to hear a dog barking instead of a human utterance.

	Unfortunately, despite various recent research efforts that apply deep generative models to produce visually aligned sound that is conditional on visual features \cite{chen2018visually, zhou2018visual}, the alignment issue remains challenging.  However, alignment is a considerably lesser concern for other sound generation tasks, including speech synthesis and voice conversion. This leads to the following research question: why is alignment especially challenging for generating visually aligned sounds?
	
	In this paper, we will uncover an important explanation from an information perspective.
	Unlike speech synthesis, where text and speech have significant information correspondence, in the task of generating sound from video,  visual and sound information do not have a strict correspondence. In most cases, sound can be decomposed into two parts. The first part is visually \textit{relevant} sound, which can be inferred from video content, \emph{e.g.}, when we see a dog open its mouth and lean forward, we will naturally associate this video clip with the sound of barking. The second part is visually \textit{irrelevant} sound that cannot be inferred from video content, \emph{e.g.}, people talking behind the camera. In the training phase of the existing sound generation paradigm, only video frames are taken as input and the loss is calculated between the predicted sound and the ground truth sound which contains both visually relevant and irrelevant information. This \textit{information mismatch} issue in the existing training paradigm may confuse the model. To minimize the loss, the model is forced to learn an incorrect mapping between video frames and visually irrelevant sound, which cripples the alignment performance.

	To generate visually aligned sound from videos, we propose a spectrogram-based sound generation model named \algnamens. Specifically, there are two mechanisms to ensure visual alignment.  First, \algname introduces a time-dependent appearance and motion feature extraction module, which provides sufficient visual information for generating the temporal and content-wise aligned sounds.
	Second and more critically, to resolve the information mismatch issue induced by visually irrelevant sound information, we introduce a novel mechanism referred to as \textit{\auxnamens}.  The \auxname passes the ground truth sound through an encoder with a bottleneck and applies the resulting bottlenecked sound features with the visual feature to predict the ground truth sound itself. We will show in the experiment part that this mechanism is helpful for visually aligned sound generation. 
	During training, the \auxname can control for the irrelevant sound component to ensure that the audio generation module can learn a correct mapping between visual features and the relevant sound component. During testing, the \auxname is removed, so that \algname is able to generate purely aligned sound only from visual feature.
	In addition, we introduce generative adversarial networks (GANs)~\cite{goodfellow2014generative} to improve the quality of the generated sound.
	We have assembled a video dataset that contains distinctive video and sound events (and thus alignment between sound and video is crucial).
	Extensive evaluations on Amazon Mechanical Turk (AMT) demonstrate that \algname significantly improves over existing methods in terms of the temporal and content-wise alignment of the generated sound.
	
	Our main contributions are listed as follows: 
	\begin{itemize}
		\item We explore the challenges of generating visually-aligned sound in the aspect of information mismatch between visual features and sounds. To solve these challenges, we propose a novel \auxname to provide missing information during the training phase.
		\item We propose \algnamens, which is an algorithm capable of generating visually aligned sound in high temporal and content-wise alignment.
	\end{itemize}

	\section{Related works}
	\textbf{Visually aligned sound synthesis.}
	Synthesizing audio for video has recently attracted considerable attention~\cite{owens2016visually,chen2017deep,zhou2018visual,goodfellow2014generative,chen2018visually}.
	Owens~\etal\cite{owens2016visually} propose the task of predicting the sound emitted by hitting and scratching objects with a drumstick.
	Chen~\etal\cite{chen2017deep} exploit conditional generative adversarial networks \cite{goodfellow2014generative} to achieve cross-modal audio-visual generation of musical performances. Chen~\etal\cite{chen2018visually} propose to generate sound considering the sound class and use perceptual loss to align semantic information.  Zhou~\etal\cite{zhou2018visual} collect an unconstraint dataset (VEGAS) that includes 10 types of sound recorded in the wild and propose a SampleRNN-based method to directly generate a waveform from a video. However, these studies disregard the information mismatch issue between video and sound, which may cause poor temporal synchronization and content-wise alignment.

	\textbf{Visual-sound learning.}
	On the one hand, deep neural networks have been widely applied for processing visual information~\cite{chen2019relation, zeng2019breaking, zeng2019graph, guo2020closed} and audio information~\cite{van2016wavenet, mehri2016samplernn} individually.
	On the other hand, the synchronized visual and sound information contained in most videos provide supervision for each other~\cite{owens2016ambient,4287000}.
	Based on visual supervision, Harwath~\etal\cite{harwath2016unsupervised} present a network to learn associations between natural image scenes and audio captions.
	Arandjelovic~\etal\cite{arandjelovic2017look} and Aytar~\etal\cite{aytar2016soundnet} learn the visual-audio correlation via unlabeled videos in an unsupervised manner.
	Several recent works~\cite{ZhaoGRVMT18,ZhaoGM019, Gan_2020_CVPR} perform visual sound separation using visual-sound representation. Also, Gan \textit{et al.}~\cite{GanZCC019} present a moving vehicle tracking system based on visual-sound relationships. 
	Some other interesting visual-sound researches include audio-visual co-segmentation~\cite{RouditchenkoZGM19} and
    audio-visual navigation~\cite{abs-1912-11684}.
	We also learn visual-sound representation and transform information from the video domain to the audio domain.
	
	\textbf{Text to speech (TTS). }
	TTS is to synthesize audio speak from text. Ling~\etal\cite{ling2015deep} use neural networks to predict sound features from pre-tokenized text features and then generate a waveform from these features. Van~\etal\cite{van2001foleyautomatic} synthesize impact sounds from physical simulations.
	Van~\etal\cite{van2016wavenet} introduce WaveNet to generate a raw audio waveform. Mehri~\etal\cite{mehri2016samplernn} propose a model for unconditional audio generation, and	Sotelo~\etal\cite{sotelo2017char2wav} present Char2Wav for speech synthesis. Shen~\etal\cite{tacotron2} propose to predict mel spectrogram from text and then transform it to a waveform using WaveNet. Our task is closely related to TTS, but we consider visual information the input and synthesize various kinds of visually aligned sounds in daily life.

	\textbf{Generative adversarial network.}
	GANs~\cite{goodfellow2014generative} have achieved considerable success in generating high-quality images~\cite{8721715,guo2019auto,8463508,8358814,8751141}. However, directly adapting image GANs architectures, such as CGANs~\cite{mirza2014conditional}, DCGANs~\cite{radford2015unsupervised} and WGANs~\cite{arjovsky2017wasserstein}, may be problematic for audio generation. Inspired by Reed~\etal\cite{reed2016generative}, which generate images conditioned on text captions,  Chen~\etal\cite{chen2017deep} design conditional GANs to achieve cross-modal audio-visual generation. By investigating waveform and spectrogram strategies, Donahue~\etal\cite{donahue2018adversarial} attempt to apply GANs to synthesize one-second slices of raw-waveform audio in an unsupervised manner. To generate high-fidelity and locally coherent audio, Engel~\etal\cite{engel2019gansynth} apply GANs to model log magnitudes and instantaneous frequencies with sufficient frequency resolution in the spectral domain. In this paper, we apply GANs on a spectrogram to improve the sound quality.
	
	\begin{figure*}[th]
		\centering
		\includegraphics[width = 1\linewidth]{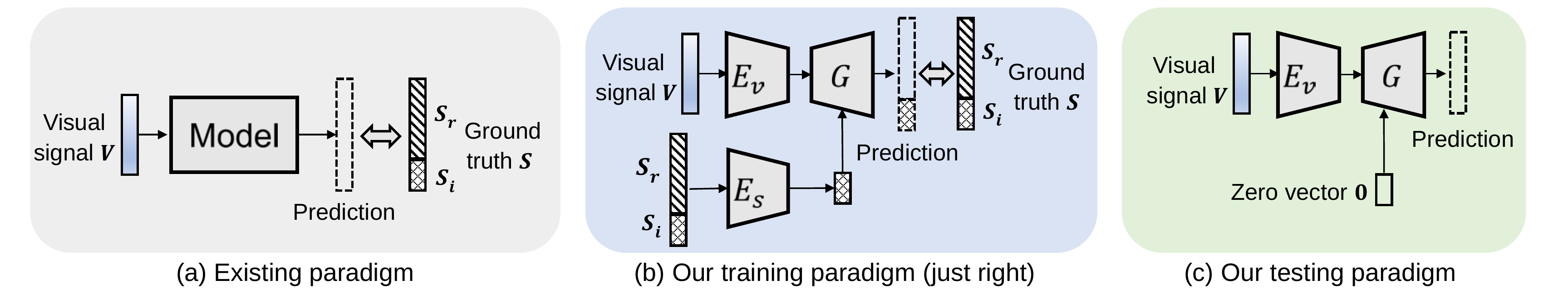}
		\caption{Comparisons between the existing paradigm and our training and testing paradigm. (a) For the existing paradigm, the model is forced to learn an incorrect mapping between a visual signal and visually irrelevant sound. (b) We avoid this situation by incorporating an \auxnamens. (c) During the testing phase, the visually relevant sound is predicted by removing the \auxnamens. }

		\label{fig:compare_paradigm}
	\end{figure*}
	
	\section{\algname framework} \label{sec:framework}
	
	In this section, we will introduce the challenge of generating visually aligned sound, together with the description about our proposed \algname framework. Then, we will discuss why the \algname is capable of solving this challenge and learn a correct mapping between video frames and visually relevant sound.
	
	\subsection{Problem formulation}
	First, we will mathematically formulate the problem and discuss its challenges.  For the remainder of this section, we will use the upper-case letters \e{X} or \e{\bm X} to denote random variables (unbolded) or random vectors (bolded), respectively, and the lower-case letters, \e{x} or \e{\bm x} to denote deterministic values. \e{\mathbb{E}[\cdot]} denotes the expectation.
	
	Denote \e{(\bm V(t), \bm S(\tau))} as a visual-sound pair. \e{\bm V(t)} represents the visual signal (vectorized) at each frame \e{t}. \e{\bm S(\tau)} represents the sound representation (waveform or spectrogram) at each frame \e{\tau}. We use different frame indexes \e{t} and \e{\tau} because visual and sound signals have different sampling rates.
	
	We introduce our formal modeling of the challenge that sound has \mr{a visually irrelevant component that can not be inferred from video content,  such as people talking behind a camera}. Assume that the audio can be decomposed into a relevant signal and an irrelevant signal:
	\begin{equation}
	\small
	\bm S(\tau) = \bm S_r(\tau) + \bm S_i(\tau),
	\end{equation}
	where the subscript \e{r} denotes ``relevant''; the subscript \e{i} denotes ``irrelevant''.
	We further assume that a relation \emph{only} exists between the video and relevant sound, which is denoted as \e{f(\cdot)}. Irrelevant sound is independent of both the relevant sound and visual features. Formally,
	\begin{equation}
	\small
	\bm S_r(\tau) = f(\bm V(t)), \quad \bm S_i(\tau) \perp \bm S_r(\tau), \quad \bm S_i(\tau) \perp \bm V(t),
	\label{eq:assump}
	\end{equation}
	where \e{\perp} denotes independence. 
	
	Our goal is to generate the visually relevant component \e{\bm S_r}~\footnote{The time argument is removed to represent that it is a collection of the signal at ALL frames.} from the visual signal \e{\bm V}. However, the visually relevant component \e{\bm S_r} is not directly available during training. Instead, we only have the sound representation \e{\bm S}, which is a mixture of \e{\bm S_r} and \e{\bm S_i}. 
	As shown in Figure~\ref{fig:compare_paradigm} (a), the existing models take the visual signal \e{\bm V} as input and regard the sound representation \e{\bm S} as target. In this setting, only the visual information is provided but the model is forced to predict both visually relevant and irrelevant sound which is independent of the visual signal. With insufficient information, the model can only learn an incorrect mapping between \e{\bm V} and \e{\bm S_i}, which produces artifacts and misalignment in generated sound. We refer to it as an information mismatch challenge.

	\begin{figure}[t]
		\centering
		\includegraphics[width = 1\linewidth]{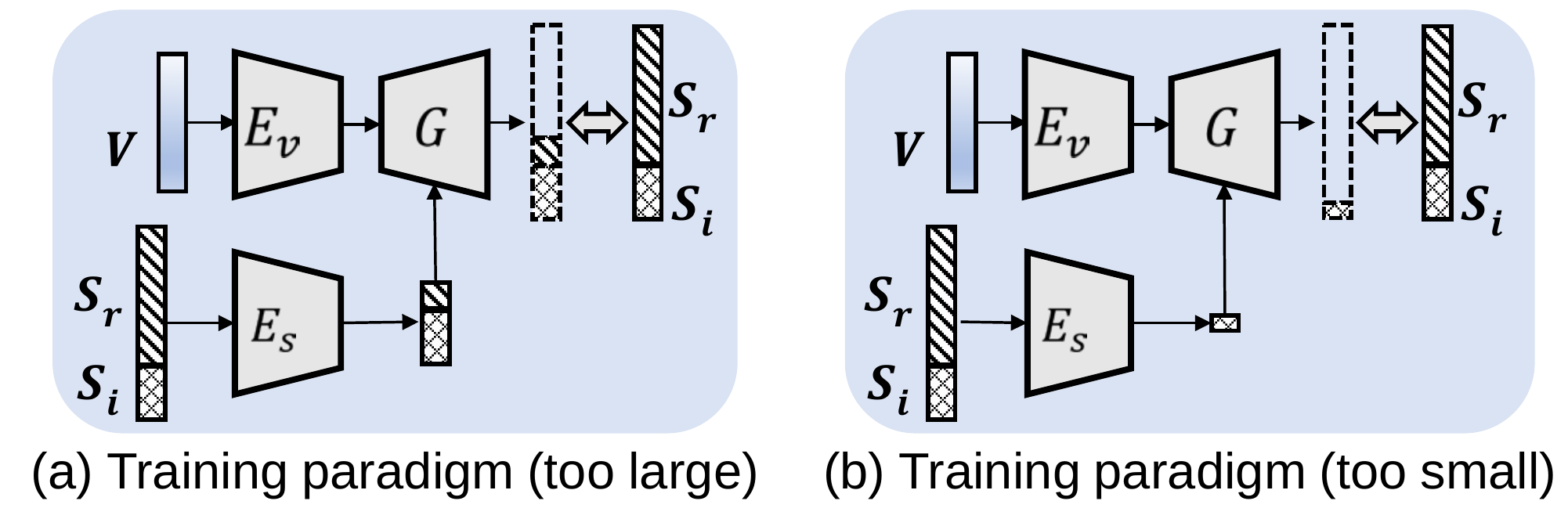}
		\caption{Influence of different information capacities of \auxnamens. (a) Because the capacity is too large, the generator perfectly reconstructs the sound but has a weak dependence on the visual signal \e{\bm V}. (b) Because the capacity is too small, the generator is forced to learn an incorrect mapping between the visual signal \e{\bm V} and the irrelevant sound \e{\bm S_i}.}
		\label{fig:variant}
	\end{figure}

	\subsection{\algname framework with audio forwarding regularizer}
	We propose a \algname to resolve the information mismatch challenge by introducing an \auxnamens. As shown in Figure~\ref{fig:compare_paradigm} (b), \algname consists of three modules: the visual encoder \e{E_v(\cdot)}, the \auxname \e{E_s(\cdot)} and the generator \e{G(\cdot, \cdot)}.  
	
	The visual encoder \e{E_v(\cdot)} takes the visual signal \e{\bm V} as the input and outputs a set of frame features. The \auxname \e{E_s(\cdot)} takes the sound signal \e{\bm S} as the input and outputs what we refer to as the regularizer output.  The generator \e{G(\cdot, \cdot)} then predicts \e{\bm S} with frame features and the regularizer output as input.  There are two different types of predictions, with or without the audio forwarding.  The prediction with audio forwarding, denoted as \e{\hat{\bm S}_a}, is predicted from both the frame features and the regularizer output. The prediction without audio forwarding, denoted as \e{\hat{\bm S}_0}, is predicted by removing \auxnamens. As shown in Figure~\ref{fig:compare_paradigm} (c), in this work, we replace the regularizer output by a zero vector when predicting \e{\hat{\bm S}_0}. Formally,
	\begin{equation}
	\small
	\hat{\bm S}_a = G(E_v(\bm V), E_s(\bm S)), \quad \hat{\bm S}_0 = G(E_v(\bm V), \bm 0).
	\label{eq:spec_gen}
	\end{equation}
	The predicted sound with audio forwarding or without audio forwarding, respectively, \emph{i.e.}, \e{\hat{\bm S}_a} and \e{\hat{\bm S}_0}, will be utilized during the training and testing phases. We exploit the adversarial training mechanism to render the predicted sound more realistic. The discriminator \e{D(\cdot, \cdot)} tries to discriminate whether the input sound is real or fake, conditional on the visual signal, \emph{i.e.}, \e{D(\bm S', \bm V)}, where \e{\bm S'} can be either generated sound or real-world sound. 
	
	From Equation \eqref{eq:spec_gen}, the significant difference between \algname and the existing paradigm is the introduction of the \auxnamens, which can be regarded an information bottleneck to control for the audio forwarding. As will be discussed in the next subsection, with careful bottleneck tuning in the training phase, \auxname can provide supplementary visually irrelevant sound information for the generator. With sufficient information in the training phase, including visual signal and irrelevant sound information, the information mismatch challenge is resolved. The generator can learn the correct mapping between a visual signal and visually aligned sound. During the testing phase, the interference of the irrelevant sound is removed, and the generator can generate visually relevant sound using only visual signals.

	\textbf{Training details:} During training, the generator tries to minimize the following loss, which involves the prediction with audio forwarding:
	\begin{equation}
	\small
	L_{{\rm rec}} + L_{{\rm G}} =
	\mathbb{E}[\lVert \hat{\bm S}_a - \bm S\rVert_2^2]
	+ \mathbb{E}[\log(1 - D(\hat{\bm S}_a,\bm V))],
	\label{eq:gen_loss}
	\end{equation}
	where the first term is the L2 reconstruction error, and the second term is the adversarial loss.
	
	On the other hand, the discriminator tries to discriminate the real sound from the predicted sound with audio forwarding, which minimizes the following standard adversarial loss:
	\begin{equation}
	\small
	L_{{\rm D}} =
	-\mathbb{E}[\log(D(\bm S, \bm V))] - \mathbb{E}[\log(1 - D(\hat{\bm S}_a, \bm V))].
	\label{eq:dis_loss}
	\end{equation}
	
	\textbf{Testing details:} In the testing phase, we will not use any regularizer output that involves the ground truth sound. Instead, we will consider the \e{\hat{\bm S}_0} defined in Equation \eqref{eq:spec_gen} the predicted results.
	
	As readers may have noticed, \algname introduces two relatively unintuitive mechanisms.  First, the \auxname sends information about \e{\bm S} to predict \e{\bm S}. This process resembles cheating during training.  Second, a mismatch exists between training and testing.  During training, \e{\hat{\bm S}_a} is employed, whereas \e{\hat{\bm S}_0} is employed during testing.  In the next subsection, we will show that these mechanisms can produce the desirable outcome.

	\begin{figure*}[th]
	\centering
	\includegraphics[width = 1\linewidth]{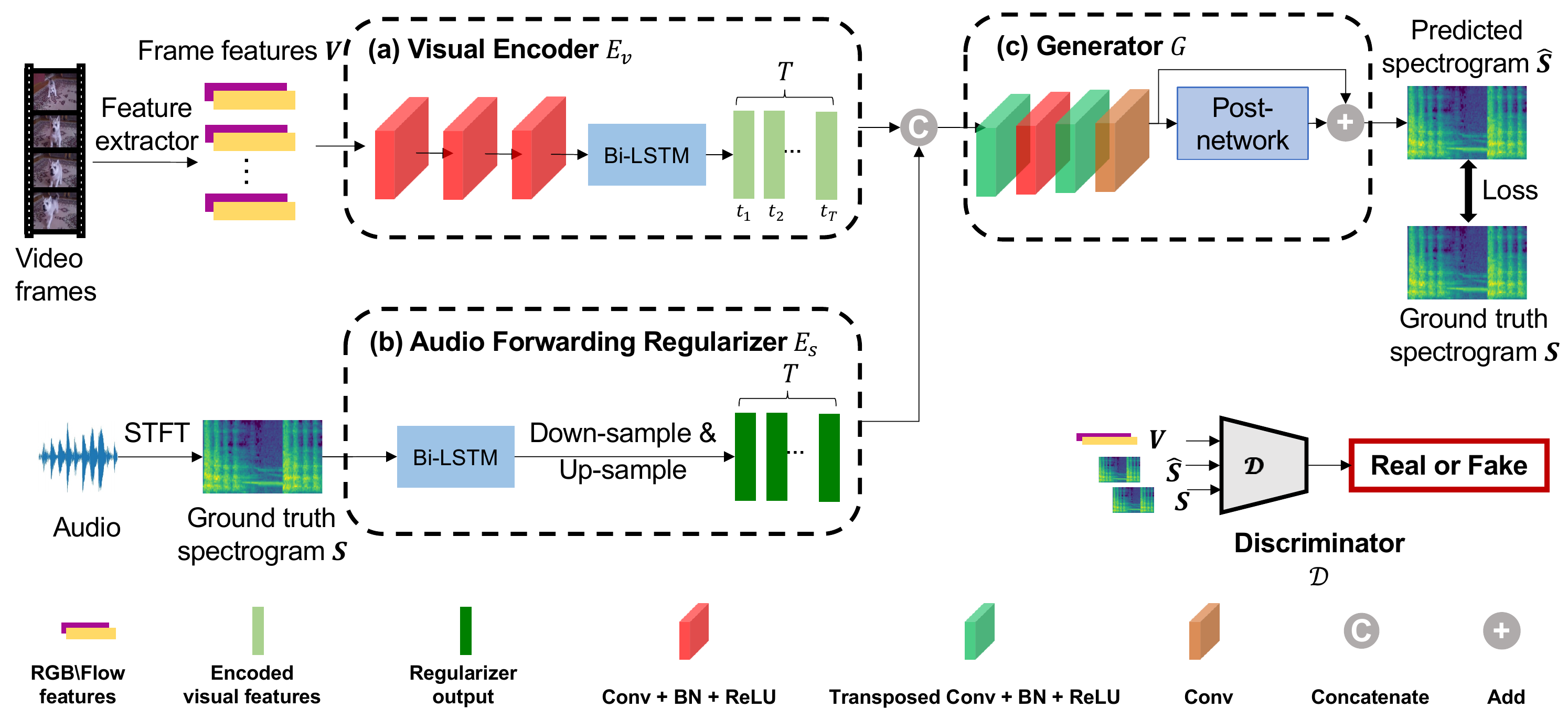}
	\caption{Schematic of our \algname. (a) The visual encoder summarizes the extracted RGB and flow frame features as time-dependent visual features. (b) The \auxname takes the real sound as input and outputs bottlenecked sound features. (c) The generator considers visual features and bottlenecked real sound information to reconstruct the real sound (spectrogram) in the training phase. }
	\label{fig:architecture}
	\end{figure*}
	
	\subsection{Why does \algname work?}
	\label{subsec:intuitive}
	In a visually sound generation task, we expect the generator to infer visually relevant sound from a visual feature. During training, however, only a mixture of visually relevant and irrelevant sounds is available to be regarded the target. The generator is forced to infer the irrelevant sound to minimize the gap between the prediction and the target. Considering the visual feature the input, the generator cannot reasonably infer irrelevant sound because they are independent of each other. To resolve this conflict, some information about visually irrelevant sound should also be considered the input. In this paper, we introduce an \auxname to provide this information. Two sources of information are available to predict \e{\bm S}: the visual feature and the regularizer output. By carefully tuning the information capacity of the audio forwarding channel, the regularizer output can take the responsibility of predicting visually irrelevant sound, which enables the generator to learn the correct mapping between a visual feature and the desired sound. In the following, we analyze the influence of information capacity in three cases.

	\textbf{Too large capacity:} if the information capacity is too large (as shown in Figure~\ref{fig:variant} (a)), then the audio forwarding channel can send all information about \e{\bm S} for prediction, which enables a perfect prediction of \e{\bm S} but a lack of dependence on the visual feature. As the capacity starts to decrease, the audio forwarding channel will prioritize sending the information about \e{\bm S_i} and enable the visual feature to address the information about \e{\bm S_r}, because the regularizer output is the only source of sound \e{\bm S_i}.
	
	\textbf{Too small capacity:} if the information capacity is too small (as shown in Figure~\ref{fig:variant} (b)), then some information about \e{\bm S_i} will not be accounted for. To minimize the reconstruction loss between the prediction and the sound \e{\bm S}, the model is forced to learn incorrect mapping between a visual feature and the irrelevant sound \e{\bm S_i}, resulting in artifacts and misalignment in the generated sound, which is the case for most existing training paradigms.
	
	\textbf{Just right capacity:} if the information capacity is just right (as shown in Figure~\ref{fig:compare_paradigm} (b)), the audio forwarding channel sends all information about \e{\bm S_i}, while the visual feature takes responsibility for the \e{\bm S_r} part. 
	The generator is allowed to learn a solid mapping between a visual feature and the relevant sound \e{\bm S_r}. During testing, the \auxname is removed and the \algname can infer the desired visually aligned sound from video content, which eliminates the inference of visually irrelevant sound.
	
	Therefore, the key to the success of \algname is to obtain proper control for the information capacity of an \auxnamens.

	\section{\algname architecture}\label{sec:arch}
	\label{sec:architecture}
	In this section, we will illustrate the architecture of each module in \algname and provide details of the vocoder, which is used to convert the predicted spectrogram into a waveform. The schematic of \algname is shown in Figure~\ref{fig:architecture}. 
	
	\textbf{Visual encoder.} We design a time-dependent visual encoder to extract appearance and motion features for visually aligned sound generation. Specifically, a BN-Inception \cite{ioffe2015batch} model is utilized as a feature extractor to explore RGB and flow features at \e{T} different time steps, which are the visual signal in Section (\ref{sec:framework}). The BN-Inception model is pretrained on ImageNet and frozen in the training phase. The concatenation of \e{T} RGB and flow features is processed by the visual encoder, which consists of three 1D convolutional layers and a two-layer bidirectional LSTM (Bi-LSTM)~\cite{hochreiter1997long}. Each convolutional layer is followed by a batch normalization (BN)~\cite{ioffe2015batch} layer and a rectified linear unit (ReLU)~\cite{nair2010rectified} activation function. 
	The outputs in the forward and backward paths of Bi-LSTM at each time step are concatenated as \e{T} encoded visual features.
	
	\textbf{Audio forwarding regularizer.}
	We design an \auxname to provide supplementary visually irrelevant sound information for sound prediction during training. The \auxname takes as input the ground truth spectrogram, which is the mixture of visually relevant and irrelevant sounds. The ground truth spectrogram is processed by a two-layer Bi-LSTM with the cell dimension \e{D}. The concatenation of the outputs in the two paths of Bi-LSTM is a \e{2D \times T'} feature map, where \e{T'} is the time dimension of the input spectrogram. Then, we uniformly downsample this feature map by \e{S}, which generates a \e{2D \times (T'/S)} feature map with some information eliminated. To match the temporal dimension of encoded visual features, we upsample it by replication and generate the \e{2D \times T} regularizer output.
	Note that the Bi-LSTM can be regarded as an information bottleneck. We control for its information capacity by changing its output dimension \e{D} and adjusting the downsampling rate \e{S}. The \auxname with a larger output dimension and smaller downsampling rate passes through richer sound information.
	The intuitive influence and empirical influence of the information capacity are discussed in subsection \ref{subsec:intuitive} and \ref{subsec:variant}, respectively.
	
	\textbf{Generator.}  
	We design a generator to predict sound (spectrogram \e{\hat{\bm S}}\footnote{We use \e{\hat{\bm S}} to represent the spectrogram with or without audio forwarding.}) from the concatenation of an encoded visual feature and regularizer output. First, we predict the initial spectrogram \e{\hat{\bm I}} by two 1D convolutional layers and two 1D transposed convolutional layers. With the exception of the last layer, each layer is followed by a BN layer and a ReLU activation function. \mr{Then, a post-network \cite{tacotron2}, which consists of five 1D convolutional layers, is introduced upon \e{\hat{\bm I}} to add the fine structure \e{\hat{ \bm R}} in a residual way. Specifically, $\hat{\bm S} = \hat{\bm I}+\hat{\bm R}$, where \e{\hat{\bm I}} and \e{\hat{ \bm R}} are the input and output of the post-network. }

	In addition to the supervision on the final predicted spectrogram $\hat{\bm S}$, we also add the L2 constraint on the initial predicted spectrogram $\hat{\bm I}$, which is expressed as follows:
	\begin{equation}
	\small
	L_{\rm rec}'=\mathbb{E}[ \lVert \hat{\bm I} - \bm S\rVert_2^2].
	\label{eq:rencon2}
	\end{equation}
	The loss in Equation (\ref{eq:gen_loss}), (\ref{eq:dis_loss}) and (\ref{eq:rencon2}) can be summarized as follows:
	\begin{equation}
	\small
	L_{total} =  L_{\rm rec} + \alpha L_{\rm rec}' + \beta (L_{\rm G}+ L_{\rm D}).
	\label{eq:total_loss}
	\end{equation}

	\textbf{Discriminator.}
	We introduce the discriminator \e{D} for adversarial training. 
	The discriminator $D$ takes as input the extracted frame feature and a spectrogram and distinguishes whether the spectrogram is derived from real video or generated by the proposed \algnamens. The PatchGANs~\cite{isola2017image} is applied to preserve the high-frequency structure in the local area. Specifically, we first separately process the input frame feature and spectrogram with two 1D transposed convolutional layers and a 1D convolutional layer. Then, we concatenate them and use four 1D convolutional layers to predict a score that indicates whether the input spectrogram is real or fake.

	\textbf{Vocoder.} Given a synthesized spectrogram, we apply WaveNet \cite{van2016wavenet} to convert it to a waveform. Instead of predicting the discretized value, we estimate a 10-component mixture of logistic distributions (MoL) to generate 16-bit samples at 22,050 Hz. We independently train the WaveNet model for each sound class.

	\section{Experiments}
	In this section, first, we describe the datasets in experiments and then provide implementation details. Second, we compare the proposed \algname with a state-of-the-art method in terms of the alignment and reality. Last, we conduct ablation studies to explore the influence from the \auxname in \algnamens.
	
	\begin{figure*}[tb]
		\centering
		\includegraphics[width = 0.9 \linewidth]{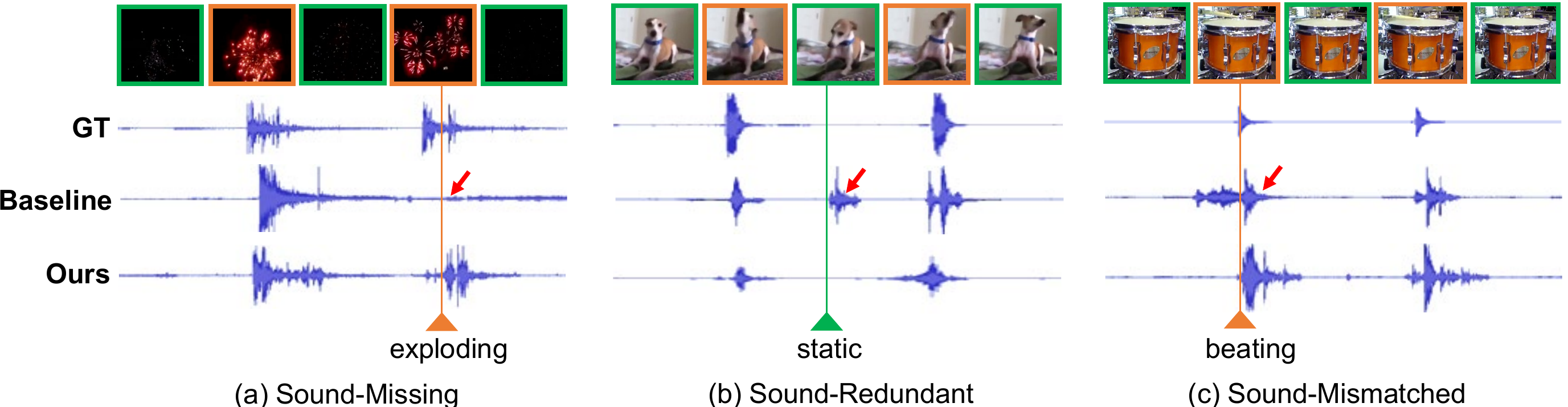}
		\caption{Evaluation on alignment with 3 misaligned cases (from left to right), including \textit{Sound-Missing}, \textit{Sound-Redundant}, and \textit{Sound-Mismatched}.}
		\label{fig:compare}
	\end{figure*}
	
	\subsection{Datasets} \label{sec:dataset}
	To evaluate the temporal alignment of our generated sound, we collect 8 video categories that have high synchronization between visual and audio contents. \mr{Such synchronization is evaluated according to the ``visual relevant'' experiment introduced in \cite{zhou2018visual}. Specifically, we exchange the sound between two videos in the same sound class and then test the turkers on Amazon Mechanical Turk (AMT) to determine whether they perceive this change. If the turkers are sensitive to this change, it means the sound is highly synchronized with the visual content. According to the results, we select  4 sound classes from  the VEGAS~\cite{zhou2018visual} dataset, namely \textbf{Fireworks, Dog, Drum, Baby crying}, and 4 sound classes from the AudioSet~\cite{audioset} dataset, namely \textbf{Cough, Hammer, Gun, Sneeze}. We denote them as \textit{Sub-VEGAS} and \textit{Sub-AudioSet} dataset respectively.
	} On average, each class contains 1626 videos, and the length for each video is 6.73 seconds. The last 128 videos from each class are selected to build a testing set.

	\subsection{Implementation details}
	
	\textbf{Data preprocessing.}
	The training videos are at 21.5 frames per second (fps) and are padded to 10 seconds by duplicating and concatenating. We extract the RGB and flow features after the last average pooling layer in the BN-Inception network. The sampling rate of audio is 22,050 Hz. The raw waveform is converted to a spectrogram via a Short Time Fourier Transform (STFT). The hop size of the STFT is set to 256 with a window length of 1,024. We transform the spectrogram on the mel scale and set the number of mel frequency bins to 80. The first 220,160 points in the raw waveform are selected for the STFT, and thus, the time dimension of the spectrogram is 860.
	
	\textbf{Training details.}
	The \algname is implemented in PyTorch. The loss is illustrated in Equation (\ref{eq:total_loss}) and we set $\alpha$ and $\beta$ to 1 and 10,000, respectively. We train the proposed \algname for 1,000 epochs and the Adam optimization method \cite{kingma2014adam} is adopted with an initial learning rate of 0.0002. Each category is independently trained. The output dimension \e{D} and downsampling rate \e{S} for the \auxname are set to 32 and 860, respectively, by default.
	
	\textbf{Baseline.} 
	Visual2Sound~\cite{zhou2018visual} is a state-of-the-art method for generating sound given visual input, which applies SampleRNN to directly generate a waveform based on the encoded visual feature. We consider it our baseline and use the generated results provided by the authors of \cite{zhou2018visual} for performance comparisons.
	
	\begin{table}[tb]
		\centering
		\caption{Evaluation of alignment. We report the percentage of the human judgments, where the results are preferred over the other methods based on three criteria. }
		\label{tab:alignment}
		\begin{tabular}{l||cc}
			\hline
			& Baseline & Ours             \\ \hline
			\textit{Sound-Missing}    & 39.38\%  & \textbf{60.62\%} \\
			\textit{Sound-Redundant}  & 40.67\%  & \textbf{59.33\%} \\
			\textit{Sound-Mismatched} & 43.23\%  & \textbf{56.77\%} \\ \hline
		\end{tabular}
	\end{table}

	\textbf{Evaluation metrics.}
	We design several metrics to evaluate the alignment and quality of the generated sound. We describe the details of each metric with its abbreviation. 1) \textbf{\textit{Sound-Missing}}: the moment when an action (or event) may reveal sound but the model fails to generate it. 2) \textbf{\textit{Sound-Redundant}}: the moment when no action (or event)  may reveal sound occurs, but the model generates redundant sound. 3) \textbf{\textit{Sound-Mismatched}}: the content of the generated sound is mismatched with the video content.  4) \textbf{\textit{Real-or-fake}}: Whether the generating sound is able to fool the human into thinking that it is a real sound. For the first three metrics, the turkers are asked to select the video with \textbf{fewer} misaligned moments, which indicates better alignment performance. For the \textit{real-or-fake} task, we present the percentage that indicates the frequency that a video is determined to be real. Each HIT is conducted by three turkers, and we aggregate their votes. The first two metrics and the third metric evaluate temporal alignment and content-wise alignment, respectively. The fourth metric measures general alignment and sound quality.

\begin{table}[tb]
	\centering
	\caption{Results of the ``real-or-fake'' task. We show the percentages that indicate the frequency that a video is determined to be real.}
	\label{tab:realfake_ave}
	
	\begin{tabular}{l||cc}
		\hline
		& Real sound & Ours             \\ \hline
		Dog       & 85.54\%    & \textbf{71.09\%} \\
		Drum      & 86.71\%    & \textbf{67.04\%} \\
		Fireworks & 91.01\%    & \textbf{86.33\%} \\
		Baby      & 91.33\%    & \textbf{72.59\%} \\
		Cough     & 82.03\%    & \textbf{55.98\%} \\
		Gun       & 91.19\%    & \textbf{70.13\%} \\
		Hammer    & 90.36\%    & \textbf{68.57\%} \\
		Sneeze    & 80.21\%    & \textbf{53.24\%} \\ \hline
		Average   & 88.65\%    & \textbf{68.12\%} \\ \hline
	\end{tabular}
\end{table}

	\subsection{Evaluation on alignment}\label{sec:evaluation_on_alignment}
	\textbf{Setup.}
	A visually aligned sound should be temporally and content-wise aligned to the video. For temporal alignment, we use \textit{Sound-Missing} and \textit{Sound-Redundant} as evaluation metric; for content-wise alignment, we use \textit{Sound-Mismatched} as evaluation metric. In the AMT test, the subjects are shown a paired video with the same visual content but different sounds, which are generated by \algname and the baseline, respectively. The subjects are required to choose the video with the better performance on alignment. The experiment is conducted on \textit{Sub-VEGAS} because the authors of \cite{zhou2018visual} only provide the results in four video types in \textit{Sub-VEGAS}.

	\textbf{Results.}
	We compare our method with baseline \cite{zhou2018visual} and report the results in Table \ref{tab:alignment}.
	\algname outperforms the baseline by 21.24\% and 18.66\% in terms of \textit{Sound-Missing} and \textit{Sound-Redundant}, respectively, which demonstrates that the proposed \algname can generate sound with better temporal alignment. Specifically, approximately 60\% of people think that the sound generated from our \algname contains less missing sound moment and less redundant sound. This finding indicates that our \algname can better capture the overlap between the visual information and the corresponding sound, even though the overlap is small. Besides, \algname outperforms baseline \cite{zhou2018visual} by 13.54\% on \textit{Sound-Mismatched}, which indicates that our \algname learns a better mapping between visual content and sound. The sound generated by \algname is better consistent with video content.

	\textbf{Qualitative visualization.}
	We visualize the waveform of real sound with the sound generated by baseline \cite{zhou2018visual} and our \algname in Fig. \ref{fig:compare}.  In the first case (a), where the fireworks are exploding, the baseline fails to generate any sound. The second case (b) shows that the baseline generates numerous redundant sounds when the dog is still. In the third case (c), where a man is hitting a drum, the baseline contains a metal hitting sound, which is not content-wise aligned. On the other hand, the sounds generated by \algname are more plausible both in temporal alignment and content-wise alignment.  We strongly encourage readers to view the demo videos in the supplementary materials.

\begin{figure}[!t]
	\includegraphics[width = 0.9 \linewidth]{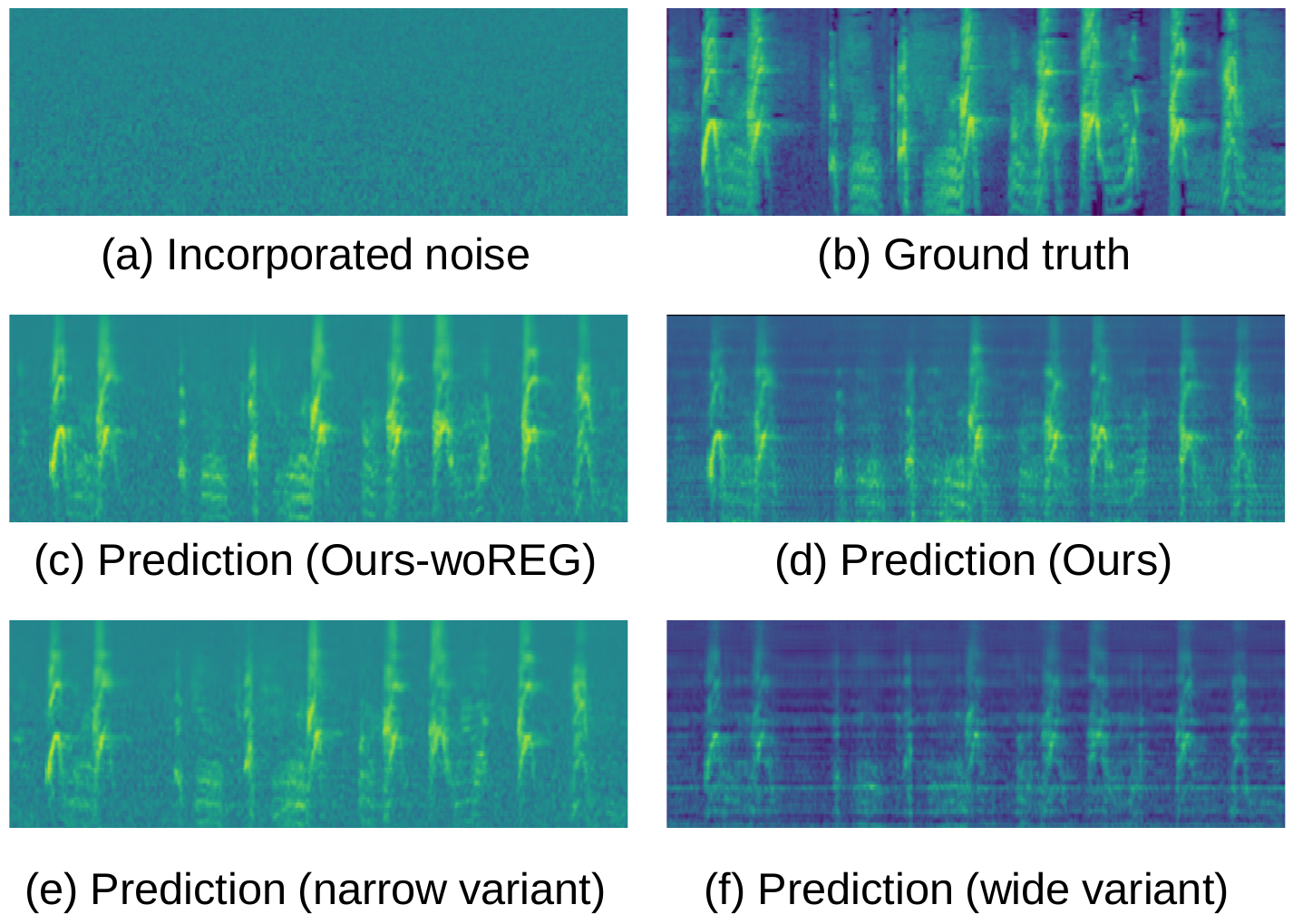}
	\caption{Spectrograms in the experiment using Dog-gauss data. We mix the Gaussian noise (a) into the ground truth audio (b) to build the Dog-gauss simulated training sample. (c), (d), (e) and (f) are the predicted spectrograms from different models, in which the one from our \algname (d) is able to get rid of the influence from the incorporated noise and preserve the sound architecture. }
	\label{fig:vis_spec}
\end{figure}

	\subsection{Evaluation on real-or-fake task}\label{subsec:naturalness}
	\textbf{Setup.}
	We conduct a ``real-or-fake'' task to evaluate whether our \algname can generate plausible sound. Specifically, we randomly combine a video with the generated sound or the real sound, and then require the turkers to make their own decisions based on the general alignment and sound quality.
	
	\textbf{Results.} From Table \ref{tab:realfake_ave}, 68\% of sounds generated from \algname success to fool human into thinking that these sounds are real, which is close to the success rate of the real sound itself. Note that the videos in the dataset have discriminative sound (as described in subsection \ref{sec:dataset}), such that the human is sensitive to the alignment of these sounds. Fooling human by the generated sound is difficult. To fool a human, the generated sound should be temporally and content-wise aligned with the visual content. Considering the Drum video for example, when the man is hitting a drum, if the model fails to generate any sound or generates a drum hitting sound soon at this time, the human may distinguish that the sound is fake. Also, if the generated sound does not sound similar to drum hitting, the human will easily perceive it.

\begin{table}[!t]
	\centering
	\caption{L2 distances ($\times\! 10^{\!-2}$) between the predicted and the ground truth spectrograms  using simulated data.}
	\label{tab:l2_distance}
	\begin{tabular}{l||ccc}
		\hline
		Training data       & Ours-woREG & Ours-VS & Ours             \\ \hline
		Dog-gauss        & 3.63  & 2.03  & \textbf{1.96} \\
		Dog-fireworks    & 3.02  & \textbf{1.82}  & 1.87 \\ \hline
	\end{tabular}
\end{table}

\begin{table}[!t]
	\centering
	\caption{Performance comparisons between \algname and the counterpart without \auxname in the ``real-or-fake'' task using simulated data.}
	\label{tab:realfake_syn}
	{
		\begin{tabular}{l||cc}
			\hline
			Training data       & Ours-woREG & Ours             \\ \hline
			Dog              & 68.75\%    & \textbf{71.09\%} (+2.34\%) \\ 
			Dog-gauss       & 58.30\%    & \textbf{67.70\%} (+9.40\%) \\
			Dog-fireworks   & 63.70\%    & \textbf{68.80\%} (+5.10\%) \\ \hline
		\end{tabular}
	}
\end{table}

	\mr{\subsection{Comparison with a simple method}}
	\mr{As we train \algname individually for each type of sound, one simple method for visually aligned sound generation is to detect the occurrence of specific actions and then play the predetermined audio clip. \aq{Specifically, we label a time step as an action occurrence if both the sound and optical flow amplitudes are higher than a threshold}. We use the visual encoder described in section~\ref{sec:architecture} to encode visual features. Then, we apply a multi-layer perceptron to perform binary classification. During testing, we manually crop several audio clips with different lengths from training data. For each detected action occurrence, we play the predetermined audio clip with a suitable length. }
	
	\mr{\aq{The “real-or-fake” task on dog video shows that only 43.65\% generated sounds can fool humans}, which is significantly worse than \algname (71.09\%). This result demonstrates that it is non-trivial to generate plausible sound. One possible reason is that such a simple method can not handle diverse sounds in one video (e.g., dog barking from young and old dogs). Besides, it is hard to define action occurrences without annotations.}

	\subsection{Effectiveness of \auxname}
	\label{subsec:effectaux}
	We aim to evaluate the performance of our \algname with and without \auxnamens. 
	\mr{We also evaluate the necessity of feeding both visual and audio information to the \auxname to learn irrelevant sound information.}

	\textbf{Simulated data.} 
	Based on the Dog data in \textit{Sub-VEGAS} dataset, we mix extra noise (visually irrelevant sound) into the original corresponding audio. With more complicated visually irrelevant sound in the training sample, it is more difficult for the model to learn the correct mapping between visual contents and visually relevant sounds. Besides, in this case, the incorporated noise becomes the main source of visually irrelevant sound, and we can approximately regard the original corresponding audio as the ground-truth visually relevant sound. Specifically, we separately incorporate two kinds of noise into the original corresponding audio, including zero-mean Gaussian noise with a standard deviation of 0.1 and the sound from fireworks videos, which generates \textbf{Dog-gauss} and \textbf{Dog-fireworks} simulated data.

\begin{table}[tb]
	\centering
	\caption{\mr{Cosine similarity between the regularizer output from Dog-fireworks sound and other sounds. The data with the same background sound generate more similar regularizer output.}}
	\label{tab:background}
	\resizebox{1\textwidth}{!}{{\color{black}
			\begin{tabular}{cc|cc}
				\hline
				\multicolumn{2}{c|}{(Similar background)} & \multicolumn{2}{c}{(Different background)} \\
				Baby-fireworks          & Dog-fireworks         & Dog-gauss               & Baby-gauss              \\ \hline
				0.48                    & 0.59                  & 0.26                    & 0.26                    \\ \hline
			\end{tabular}
	}}
\end{table}

\begin{table}[tb]
	\centering
	\caption{\mr{L2 distances ($\times\! 10^{\!-2}$) on spectrogram to visually relevant and irrelevant sounds when setting visual feature to zero and feeding input data to regularizer only. (The irrelevant sounds indicate  Gaussian, fireworks and drum sounds, respectively for three different input data.)}}
	\label{tab:spec}
	\resizebox{1\textwidth}{!}{{\color{black}
			\begin{tabular}{l||cc}
				\hline
				\multicolumn{1}{l||}{\multirow{2}{*}{Input data}} & Dog              & Gauss. or fireworks or drum \\
				\multicolumn{1}{l||}{}                            & (relevant sound) & (irrelevant sound)        \\ \hline
				Dog-gauss                                        & 4.36             & \textbf{0.84}                     \\
				Dog-fireworks                                    & 3.69             & \textbf{1.97}                      \\ 
				\color{black}{Dog-drum}                                    & \color{black}{4.53}             & \color{black}{\textbf{2.29}}                      \\ \hline			\end{tabular}
	}}
\end{table}

	\textbf{Setup.}
	In addition to the proposed \algnamens, we construct a counterpart in which we remove the \auxname in both the training pahse and testing phase, and refer to it as \textbf{Ours-woREG}. 
	\mr{Also, we input both visual feature and spectrogram into the \auxname to construct a variant, namely \textbf{Ours-VS}.}
	We train these models using simulated data to evaluate whether our \algname is robust to different kinds of visually irrelevant sound in the training samples. We report the success rate for a ``real-or-fake'' task because this metric is sensitive to the quality of the generated sound, especially the alignment and reality. Besides, considering the original audio in the simulated data as ground truth, we use the L2 distance to evaluate the performance of \algname in generating visually relevant sound.
	
\begin{table}[t]
	\centering
	\caption{\mr{L2 distances ($\times\! 10^{\!-2}$) between the predicted and the background sound spectrograms when feeding into zero vector or regularizer output. (The regularizer output indicates the regularizer output from Dog-gauss, Dog-fireworks and Dog-drum, respectively when comparing with different background sound spectrogram.)}}
	\label{tab:mixin}
	{\color{black}
		\begin{tabular}{l||cc}
			\hline
			Background sound & Zero vector & Regularizer output  \\ \hline
			Gaussian   & 5.05        & $\textbf{2.72}$ \\
			Fireworks  & 5.56     & $\textbf{2.78}$ \\ 
			\color{black}{Drum}  & \color{black}{6.11}     & \color{black}{$\textbf{2.79}$} \\ \hline
		\end{tabular}
	}
\end{table}

	\begin{table*}[tb]
		\centering
		\caption{Performance of \algname variants. The L2 distance ($\times\! 10^{\!-2}$) is evaluated on the Dog-gauss data, while the other metrics are evaluated on both \textit{Sub-VEGAS} data and \textit{Sub-AudioSet} data.}
		\label{tab:variant}
		\begin{tabular}{ll||c|cccc}
			\hline
			&                   & L2 distance   & \textit{real-or-fake} & \textit{Sound-Missing} & \textit{Sound-Redundant} & \textit{Sound-Mismatched} \\
			&                   & (Dog-gauss)   & \multicolumn{4}{c}{(\textit{Sub-VEGAS} and \textit{Sub-AudioSet})}                                                     \\ \hline
			Narrow variant                & $S_{860}D_{8}$    & 3.18          & 70.18\%               & 3.16                   & 3.01                     & 2.95                      \\
			\multirow{2}{*}{Wide variant} & $S_{860}D_{1024}$ & 3.19          & 62.50\%               & 2.96                   & 3.20                     & 2.92                      \\
			& $S_{32}D_{32}$    & 3.08          & 66.67\%               & 2.94                   & 3.22                     & 2.93                      \\
			\algnamens                    & $S_{860}D_{32}$   & \textbf{1.96} & \textbf{71.09\%} & \textbf{3.20}          & \textbf{3.35}            & \textbf{3.02}             \\ \hline
		\end{tabular}

	\end{table*}

	\textbf{Objective results.}
	In Table \ref{tab:l2_distance}, we show the L2 distance between the predicted spectrogram and the ground truth spectrogram in the experiments using simulated data. Note that we consider the original audio the ground truth.
	With two kinds of simulated data, the predicted spectrogram from our \algname and Ours-VS is closer to the ground truth compared with Ours-woREG counterpart (1.96, 2.03 v.s. 3.63 and 1.87, 1.82 v.s. 3.02, respectively). This finding indicates that the \auxname enables \algname to perform better in generating visually relevant sound. \mr{Also, it demonstrates that the \auxname does not need to take as input visual content explicitly. We argue that because the \auxname is trained for each type of video individually, the video category information is available for the network as one kind of visual information. Such category information is enough for regularizer to learn visually irrelevant sound. }
	In Fig. \ref{fig:vis_spec}, we visualize four spectrograms, including the incorporated Gaussian noise, original audio (ground truth), and predicted spectrograms from our \algname and Ours-woREG counterpart. The Ours-woREG counterpart has learned some incorrect mappings between a visual feature and visually irrelevant noise, while our \algname focuses only on generating the visually relevant component.

	\textbf{Subjective results.}
	In Table \ref{tab:realfake_syn}, we show the performance of our \algname and Ours-woREG counterpart in the ``real-or-fake'' task. Our \algname consistently outperforms the Our-woREG counterpart, and achieves performance gains of 2.34\%, 9.40\%, and 5.10\% on three training data. This finding demonstrates that the \auxname is important for generating  realistic sound. Note that the incorporated irrelevant sounds (\emph{i.e.}, Gaussian noise and fireworks sound) dramatically deteriorate the performance and cause a 10.45\% and 5.05\% decrease in performance, while our \algname is robust to the irrelevant sounds. One possible reason is that the proposed \auxname is helpful for eliminating the negative influence from irrelevant sounds in the training phase, which enables the model to learn proper mapping between visual features and visually relevant sounds.

	\mr{\subsection{Exploration on regularizer output}}
\mr{We aim to explore what the \auxname has learned. Specifically, we will answer the following three questions: 1) is the regularizer output from spectrograms with similar background sounds more similar than those with different background sounds? 2) what information can be reconstructed from regularizer output? 3) can the generator mix in sounds from regularizer output during testing?}

\mr{\textbf{Setup.} }
\aq{To answer these questions, we generate several types of simulated data, namely Dog-gauss, Dog-fireworks, Dog-drum, Baby-gauss, and Baby-fireworks. The term behind the hyphen (-) can be regarded as visually irrelevant sound. Each type of simulated data includes 128 videos for testing.}

\aq{For the first question, we train four audio forwarding regularizers individually on Dog-fireworks, Dog-gauss, Baby-fireworks, and Baby-gauss data. Then, we input these four types of sound to the corresponding trained \auxname to generate four types of regularizer outputs.}
We use cosine distance to measure the similarity between regularizer outputs generated from different sounds.

\aq{For the second question, we study what information can the model reconstruct when it only takes regularizer output as input. Specifically, we train three \algname individually on Dog-gauss, Dog-fireworks, and Dog-drum data. For each trained \algnamens, we set the visual feature as zero and input the corresponding regularizer output to generate spectrogram. We measure the similarity between the generated spectrogram and visually relevant (or irrelevant) sound.}

\aq{For the third question, we train a \algname on dog videos and try to mix in background sounds (\textit{i.e.}, Gaussian, fireworks, or drum)  by replacing zero vector with different representations (\textit{i.e.}, the regularizer output from Dog-gauss, Dog-fireworks, or Dog-drum).} To measure whether the generated spectrogram include background sound, we compute the L2 distance between spectrogram of generated and background sounds.

\mr{\textbf{Results.} Table \ref{tab:background} shows the cosine similarity between regularizer output from Dog-fireworks and other simulated data.  The data with fireworks background sound (\textit{i.e.}, Baby-fireworks and Dog-fireworks) have higher similarity values compared with those with other background sounds. This indicates the regularizer passes through the background sound information ignoring the visually relevant sound. 
	For the second question, in Table \ref{tab:spec}, the generated spectrogram have smaller L2 distances to visually irrelevant sound compared to relevant sound (0.84 \textit{v.s} 4.36 and 1.97 \textit{v.s} 3.69 and 2.29 \textit{v.s} 4.53). This indicates the regularizer is good at reconstructing the irrelevant sound component (\textit{i.e.}, Gaussian, fireworks and drum sounds) compared to the relevant part (\textit{i.e.}, dog sound). We believe the reason is that the regularizer output mainly contains irrelevant sound information instead of the relevant one.
	For the third question, the L2 distances between the spectrogram of generated and background sounds are shown in Table \ref{tab:mixin}.
	When we feed in regularizer output instead of zero vector, the distances are relatively smaller (2.72 \textit{v.s} 5.05 and 2.78 \textit{v.s} 5.56 and 2.79 \textit{v.s} 6.11), indicating that the background sound information contained in regularizer output is successfully mixed in the output.
}

\vspace{0.5cm}
	\subsection{Exploration on \algname variants}
	\label{subsec:variant}
	As intuitively explained in subsection~\ref{subsec:intuitive}, with appropriate control for the capacity of the \auxnamens, the \algname can predict visually relevant sound. Next, we will show how we control for the capacity of the \auxname and how it affects the prediction \e{\hat{S_0}}.
	
	\textbf{\algname variants.} We control for the capacity of the \auxname by changing its downsampling rate \e{S} and output dimension \e{D}. By default, \e{S} and \e{D} are set to 860 and 32 in our \algname, and thus, we denote it as \e{S_{860}D_{32}}. With a larger output dimension or by sampling the regularizer output more densely, more information can be passed through the \auxnamens.  Thus, we design two wide \algname variants \e{S_{860}D_{1024}} and \e{S_{32}D_{32}}, whose \auxname has a larger output dimension of 1024 or a smaller downsampling rate of 32. Conversely, we decrease \e{D} to 8 and construct the narrow variant \e{S_{860}D_{8}}.

	\textbf{Setup.}
	We train the \algname variants with different capacities on Dog-gauss data and leverage the L2 distance to evaluate their performance in generating visually relevant sound. Besides, we conduct experiments on \textit{Sub-VEGAS} and \textit{Sub-AudioSet} and leverage the ``real-or-fake'' task with \textit{Sound-Missing}, \textit{Sound-Redundant} and \textit{Sound-Mismatched} metrics to evaluate the quality and alignment of the generated sound from different \algname variants. To make comparisons among several variants, the three alignment metrics are evaluated by the mean opinion score (MOS) test. Specifically, we provide two reference videos for the turkers and score them as 1 and 5. The higher score indicates better alignment. The turkers are asked to assign a score of 1-5 according to the corresponding requirements.

	\textbf{Objective results.}
	In Table \ref{tab:variant}, we show the L2 distance in the experiment with Dog-gauss data. Compared with our \algnamens, both the ``narrow variant'' and the ``wide variant'' have a larger L2 distance between the prediction and the ground truth sound (1.96 v.s. 3.18, 3.19, 3.08). This finding indicates that the optimal results are achieved only when the capacity of the \auxname is appropriately controlled. We further visualize the predicted spectrogram \e{\hat{S_0}} in Fig. \ref{fig:vis_spec}. For the ``wide variant'', when the \auxname is shut down, the generator misses some of the meaningful spectrogram structure. This is not surprising because the \auxname with a large capacity is able to provide sufficient information about the target spectrogram during the training phase. The generator tends to use this information for spectrogram reconstruction instead of learning mapping between a visual feature and visually relevant sound. For the ``narrow variant'', the predicted spectrogram contains considerable noise because the model is forced to learn some incorrect mapping between a visual feature and noise.
	
	\textbf{Subjective results.}
	In Table \ref{tab:variant}, our \algname performs the best for both the ``real-or-fake'' task and three alignment metrics. Specifically, 71.09\% generated sounds from \algname are considered real sounds, compared with 62.50\% and 66.67\% generated sounds for the ``wide variant'' and 70.18\% for the ``narrow variant''. Our \algname also achieves the highest scores with all three alignment metrics. Note that the ``wide variant'' achieves low scores for the \textit{Sound-Missing} term, which demonstrates that the model fails to generate sound when action occurs. The reason is that the \auxname with excessively large capacity prevents the model from learning necessary mapping between a visual feature and sound.


	\mr{\subsection{Exploration on discriminator}}
	\mr{As there are multiple possible sounds that are visually aligned with the visual content, L2 reconstruction loss can only supervise the network to learn an overall sound structure. This may result in a smooth spectrogram and low-fidelity results. Adversarial training mechanism has been proven to be effective to generate high-fidelity and coherent audio in several works \cite{engel2019gansynth,donahue2018adversarial,chen2017deep}. To study the necessity of discriminator in our \algnamens, we train a \algname without adversarial loss and conduct a ``real-or-fake'' human study on \textit{Sub-VEGAS} and \textit{Sub-AudioSet} datasets. Ours-woGAN fools 64.04\% human on average, which is worse than \algname (68.12\%). This result demonstrates that the discriminator is helpful to generate high-fidelity sounds in \algnamens.}

	\section{Conclusion}
	We explain from an information mismatch perspective why the temporal and content-wise alignment remains challenging for generating sound from videos, and propose an \auxname to solve this challenge. With the \auxnamens, our \algname can leverage sufficient information for prediction during the training phase and learn a concrete correspondence between visual information and sound information. Experiments show that the sound generated by our \algname is more plausible and more aligned to  video.

	\ifCLASSOPTIONcaptionsoff
	\newpage
	\fi

	\bibliographystyle{IEEEtran}
	\bibliography{IEEEabrv,video2sound}

\begin{thebibliography}{10}
\providecommand{\url}[1]{#1}
\csname url@samestyle\endcsname
\providecommand{\newblock}{\relax}
\providecommand{\bibinfo}[2]{#2}
\providecommand{\BIBentrySTDinterwordspacing}{\spaceskip=0pt\relax}
\providecommand{\BIBentryALTinterwordstretchfactor}{4}
\providecommand{\BIBentryALTinterwordspacing}{\spaceskip=\fontdimen2\font plus
\BIBentryALTinterwordstretchfactor\fontdimen3\font minus
  \fontdimen4\font\relax}
\providecommand{\BIBforeignlanguage}[2]{{%
\expandafter\ifx\csname l@#1\endcsname\relax
\typeout{** WARNING: IEEEtran.bst: No hyphenation pattern has been}%
\typeout{** loaded for the language `#1'. Using the pattern for}%
\typeout{** the default language instead.}%
\else
\language=\csname l@#1\endcsname
\fi
#2}}
\providecommand{\BIBdecl}{\relax}
\BIBdecl

\bibitem{chen2018visually}
K.~Chen, C.~Zhang, C.~Fang, Z.~Wang, T.~Bui, and R.~Nevatia, ``Visually
  indicated sound generation by perceptually optimized classification,'' in
  \emph{The European Conference on Computer Vision}, 2018, pp. 560--574.

\bibitem{zhou2018visual}
Y.~Zhou, Z.~Wang, C.~Fang, T.~Bui, and T.~L. Berg, ``Visual to sound:
  Generating natural sound for videos in the wild,'' in \emph{The IEEE
  Conference on Computer Vision and Pattern Recognition}, 2018, pp. 3550--3558.

\bibitem{goodfellow2014generative}
I.~J. Goodfellow, J.~Pouget{-}Abadie, M.~Mirza, B.~Xu, D.~Warde{-}Farley,
  S.~Ozair, A.~C. Courville, and Y.~Bengio, ``Generative adversarial nets,'' in
  \emph{Advances in Neural Information Processing Systems}, 2014, pp.
  2672--2680.

\bibitem{owens2016visually}
A.~Owens, P.~Isola, J.~McDermott, A.~Torralba, E.~H. Adelson, and W.~T.
  Freeman, ``Visually indicated sounds,'' in \emph{Proceedings of the IEEE
  conference on computer vision and pattern recognition}, 2016, pp. 2405--2413.

\bibitem{chen2017deep}
L.~Chen, S.~Srivastava, Z.~Duan, and C.~Xu, ``Deep cross-modal audio-visual
  generation,'' in \emph{Proceedings of the on Thematic Workshops of ACM
  Multimedia}, 2017, pp. 349--357.

\bibitem{chen2019relation}
P.~Chen, C.~Gan, G.~Shen, W.~Huang, R.~Zeng, and M.~Tan, ``Relation attention
  for temporal action localization,'' \emph{IEEE Transactions on Multimedia},
  2019.

\bibitem{zeng2019breaking}
R.~Zeng, C.~Gan, P.~Chen, W.~Huang, Q.~Wu, and M.~Tan, ``Breaking
  winner-takes-all: Iterative-winners-out networks for weakly supervised
  temporal action localization,'' \emph{IEEE Transactions on Image Processing},
  vol.~28, no.~12, pp. 5797--5808, 2019.

\bibitem{zeng2019graph}
R.~Zeng, W.~Huang, M.~Tan, Y.~Rong, P.~Zhao, J.~Huang, and C.~Gan, ``Graph
  convolutional networks for temporal action localization,'' in \emph{The IEEE
  International Conference on Computer Vision}, Oct 2019.

\bibitem{guo2020closed}
Y.~Guo, J.~Chen, J.~Wang, Q.~Chen, J.~Cao, Z.~Deng, Y.~Xu, and M.~Tan,
  ``Closed-loop matters: Dual regression networks for single image
  super-resolution,'' in \emph{Proceedings of the IEEE Conference on Computer
  Vision and Pattern Recognition}, 2020.

\bibitem{van2016wavenet}
A.~van~den Oord, S.~Dieleman, H.~Zen, K.~Simonyan, O.~Vinyals, A.~Graves,
  N.~Kalchbrenner, A.~W. Senior, and K.~Kavukcuoglu, ``Wavenet: {A} generative
  model for raw audio,'' in \emph{ISCA Speech Synthesis Workshop}, 2016, p.
  125.

\bibitem{mehri2016samplernn}
S.~Mehri, K.~Kumar, I.~Gulrajani, R.~Kumar, S.~Jain, J.~Sotelo, A.~C.
  Courville, and Y.~Bengio, ``{SampleRNN}: An unconditional end-to-end neural
  audio generation model,'' in \emph{International Conference on Learning
  Representations}, 2017.

\bibitem{owens2016ambient}
A.~Owens, J.~Wu, J.~H. McDermott, W.~T. Freeman, and A.~Torralba, ``Ambient
  sound provides supervision for visual learning,'' in \emph{The European
  Conference on Computer Vision}, 2016, pp. 801--816.

\bibitem{4287000}
G.~{Monaci}, P.~{Jost}, P.~{Vandergheynst}, B.~{Mailhe}, S.~{Lesage}, and
  R.~{Gribonval}, ``Learning multimodal dictionaries,'' \emph{IEEE Transactions
  on Image Processing}, vol.~16, no.~9, pp. 2272--2283, 2007.

\bibitem{harwath2016unsupervised}
D.~F. Harwath, A.~Torralba, and J.~R. Glass, ``Unsupervised learning of spoken
  language with visual context,'' in \emph{Advances in Neural Information
  Processing Systems}, 2016, pp. 1858--1866.

\bibitem{arandjelovic2017look}
R.~Arandjelovic and A.~Zisserman, ``Look, listen and learn,'' in \emph{The IEEE
  International Conference on Computer Vision}, 2017, pp. 609--617.

\bibitem{aytar2016soundnet}
Y.~Aytar, C.~Vondrick, and A.~Torralba, ``Soundnet: Learning sound
  representations from unlabeled video,'' in \emph{Advances in Neural
  Information Processing Systems}, 2016, pp. 892--900.

\bibitem{ZhaoGRVMT18}
H.~Zhao, C.~Gan, A.~Rouditchenko, C.~Vondrick, J.~H. McDermott, and
  A.~Torralba, ``The sound of pixels,'' in \emph{European conference on
  computer vision}, 2018, pp. 587--604.

\bibitem{ZhaoGM019}
H.~Zhao, C.~Gan, W.~Ma, and A.~Torralba, ``The sound of motions,'' in
  \emph{{IEEE} International Conference on Computer Vision}, 2019, pp.
  1735--1744.

\bibitem{Gan_2020_CVPR}
C.~Gan, D.~Huang, H.~Zhao, J.~B. Tenenbaum, and A.~Torralba, ``Music gesture
  for visual sound separation,'' in \emph{{IEEE} Conference on Computer Vision
  and Pattern Recognition}, 2020.

\bibitem{GanZCC019}
C.~Gan, H.~Zhao, P.~Chen, D.~Cox, and A.~Torralba, ``Self-supervised moving
  vehicle tracking with stereo sound,'' in \emph{{IEEE} International
  Conference on Computer Vision}, 2019, pp. 7052--7061.

\bibitem{RouditchenkoZGM19}
A.~Rouditchenko, H.~Zhao, C.~Gan, J.~H. McDermott, and A.~Torralba,
  ``Self-supervised audio-visual co-segmentation,'' in \emph{{IEEE}
  International Conference on Acoustics, Speech and Signal Processing}, 2019,
  pp. 2357--2361.

\bibitem{abs-1912-11684}
C.~Gan, Y.~Zhang, J.~Wu, B.~Gong, and J.~B. Tenenbaum, ``Look, listen, and act:
  Towards audio-visual embodied navigation,'' \emph{IEEE International
  Conference on Robotics and Automation}, 2020.

\bibitem{ling2015deep}
Z.~Ling, S.~Kang, H.~Zen, A.~W. Senior, M.~Schuster, X.~Qian, H.~M. Meng, and
  L.~Deng, ``Deep learning for acoustic modeling in parametric speech
  generation: {A} systematic review of existing techniques and future trends,''
  \emph{IEEE Signal Processing Magazine}, vol.~32, no.~3, pp. 35--52, 2015.

\bibitem{van2001foleyautomatic}
K.~van~den Doel, P.~G. Kry, and D.~K. Pai, ``{FoleyAutomatic}: physically-based
  sound effects for interactive simulation and animation,'' in
  \emph{Proceedings of the Annual Conference on Computer Graphics and
  Interactive Techniques}, 2001, pp. 537--544.

\bibitem{sotelo2017char2wav}
J.~Sotelo, S.~Mehri, K.~Kumar, J.~F. Santos, K.~Kastner, A.~C. Courville, and
  Y.~Bengio, ``{Char2Wav}: End-to-end speech synthesis,'' in
  \emph{International Conference on Learning Representations}, 2017.

\bibitem{tacotron2}
J.~Shen, R.~Pang, R.~J. Weiss, M.~Schuster, N.~Jaitly, Z.~Yang, Z.~Chen,
  Y.~Zhang, Y.~Wang, R.~Ryan, R.~A. Saurous, Y.~Agiomyrgiannakis, and Y.~Wu,
  ``Natural {TTS} synthesis by conditioning wavenet on {MEL} spectrogram
  predictions,'' in \emph{IEEE International Conference on Acoustics, Speech
  and Signal Processing}, 2018, pp. 4779--4783.

\bibitem{8721715}
F.~{Yu}, X.~{Wu}, J.~{Chen}, and L.~{Duan}, ``Exploiting images for video
  recognition: Heterogeneous feature augmentation via symmetric adversarial
  learning,'' \emph{IEEE Transactions on Image Processing}, vol.~28, no.~11,
  pp. 5308--5321, 2019.

\bibitem{guo2019auto}
Y.~Guo, Q.~Chen, J.~Chen, Q.~Wu, Q.~Shi, and M.~Tan, ``Auto-embedding
  generative adversarial networks for high resolution image synthesis,''
  \emph{{IEEE Transactions on Multimedia}}, vol.~21, pp. 2726--2737, 2019.

\bibitem{8463508}
X.~{Chen}, C.~{Xu}, X.~{Yang}, L.~{Song}, and D.~{Tao}, ``{Gated-GAN}:
  Adversarial gated networks for multi-collection style transfer,'' \emph{IEEE
  Transactions on Image Processing}, vol.~28, no.~2, pp. 546--560, 2019.

\bibitem{8358814}
C.~{Wang}, C.~{Xu}, C.~{Wang}, and D.~{Tao}, ``Perceptual adversarial networks
  for image-to-image transformation,'' \emph{IEEE Transactions on Image
  Processing}, vol.~27, no.~8, pp. 4066--4079, 2018.

\bibitem{8751141}
C.~{Hsu}, C.~{Lin}, W.~{Su}, and G.~{Cheung}, ``{SiGAN}: Siamese generative
  adversarial network for identity-preserving face hallucination,'' \emph{IEEE
  Transactions on Image Processing}, vol.~28, no.~12, pp. 6225--6236, 2019.

\bibitem{mirza2014conditional}
M.~Mirza and S.~Osindero, ``Conditional generative adversarial nets,''
  \emph{arXiv preprint arXiv:1411.1784}, 2014.

\bibitem{radford2015unsupervised}
A.~Radford, L.~Metz, and S.~Chintala, ``Unsupervised representation learning
  with deep convolutional generative adversarial networks,'' in
  \emph{International Conference on Learning Representations}, 2016.

\bibitem{arjovsky2017wasserstein}
M.~Arjovsky, S.~Chintala, and L.~Bottou, ``Wasserstein generative adversarial
  networks,'' in \emph{Proceedings of the International Conference on Machine
  Learning}, 2017, pp. 214--223.

\bibitem{reed2016generative}
S.~E. Reed, Z.~Akata, X.~Yan, L.~Logeswaran, B.~Schiele, and H.~Lee,
  ``Generative adversarial text to image synthesis,'' in \emph{Proceedings of
  the International Conference on Machine Learning}, 2016, pp. 1060--1069.

\bibitem{donahue2018adversarial}
C.~Donahue, J.~McAuley, and M.~Puckette, ``Adversarial audio synthesis,'' in
  \emph{International Conference on Learning Representations}, 2019.

\bibitem{engel2019gansynth}
J.~Engel, K.~K. Agrawal, S.~Chen, I.~Gulrajani, C.~Donahue, and A.~Roberts,
  ``{GANS}ynth: Adversarial neural audio synthesis,'' in \emph{International
  Conference on Learning Representations}, 2019.

\bibitem{ioffe2015batch}
S.~Ioffe and C.~Szegedy, ``Batch normalization: Accelerating deep network
  training by reducing internal covariate shift,'' in \emph{International
  Conference on Learning Representations}, 2015, pp. 448--456.

\bibitem{hochreiter1997long}
S.~Hochreiter and J.~Schmidhuber, ``Long short-term memory,'' \emph{Neural
  computation}, vol.~9, no.~8, pp. 1735--1780, 1997.

\bibitem{nair2010rectified}
V.~Nair and G.~E. Hinton, ``Rectified linear units improve restricted boltzmann
  machines,'' in \emph{{ICML}}, 2010, pp. 807--814.

\bibitem{isola2017image}
P.~Isola, J.~Zhu, T.~Zhou, and A.~A. Efros, ``Image-to-image translation with
  conditional adversarial networks,'' in \emph{The IEEE Conference on Computer
  Vision and Pattern Recognition}, 2017, pp. 5967--5976.

\bibitem{audioset}
J.~F. Gemmeke, D.~P. Ellis, D.~Freedman, A.~Jansen, W.~Lawrence, R.~C. Moore,
  M.~Plakal, and M.~Ritter, ``Audio set: An ontology and human-labeled dataset
  for audio events,'' in \emph{The IEEE International Conference on Acoustics,
  Speech and Signal Processing}, 2017, pp. 776--780.

\bibitem{kingma2014adam}
D.~P. Kingma and J.~Ba, ``Adam: {A} method for stochastic optimization,'' in
  \emph{International Conference on Learning Representations}, 2015.

\end{thebibliography}

	\begin{IEEEbiography}[{\includegraphics[width=1in,height=1.25in,clip,keepaspectratio]{{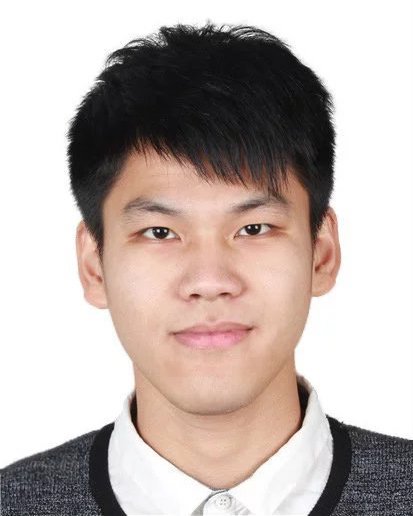}}}]{Peihao Chen}
	received the B.E. degree in Automation Science and Engineering from South China University of Technology, China, in 2018. He is working toward the M.E. degree in the School of Software Engineering,
	South China University of Technology, China. His research interests include Deep Learning in Video and Audio Understanding.
	\end{IEEEbiography}
	
	\begin{IEEEbiography}[{\includegraphics[width=1in,height=1.25in,clip,keepaspectratio]{{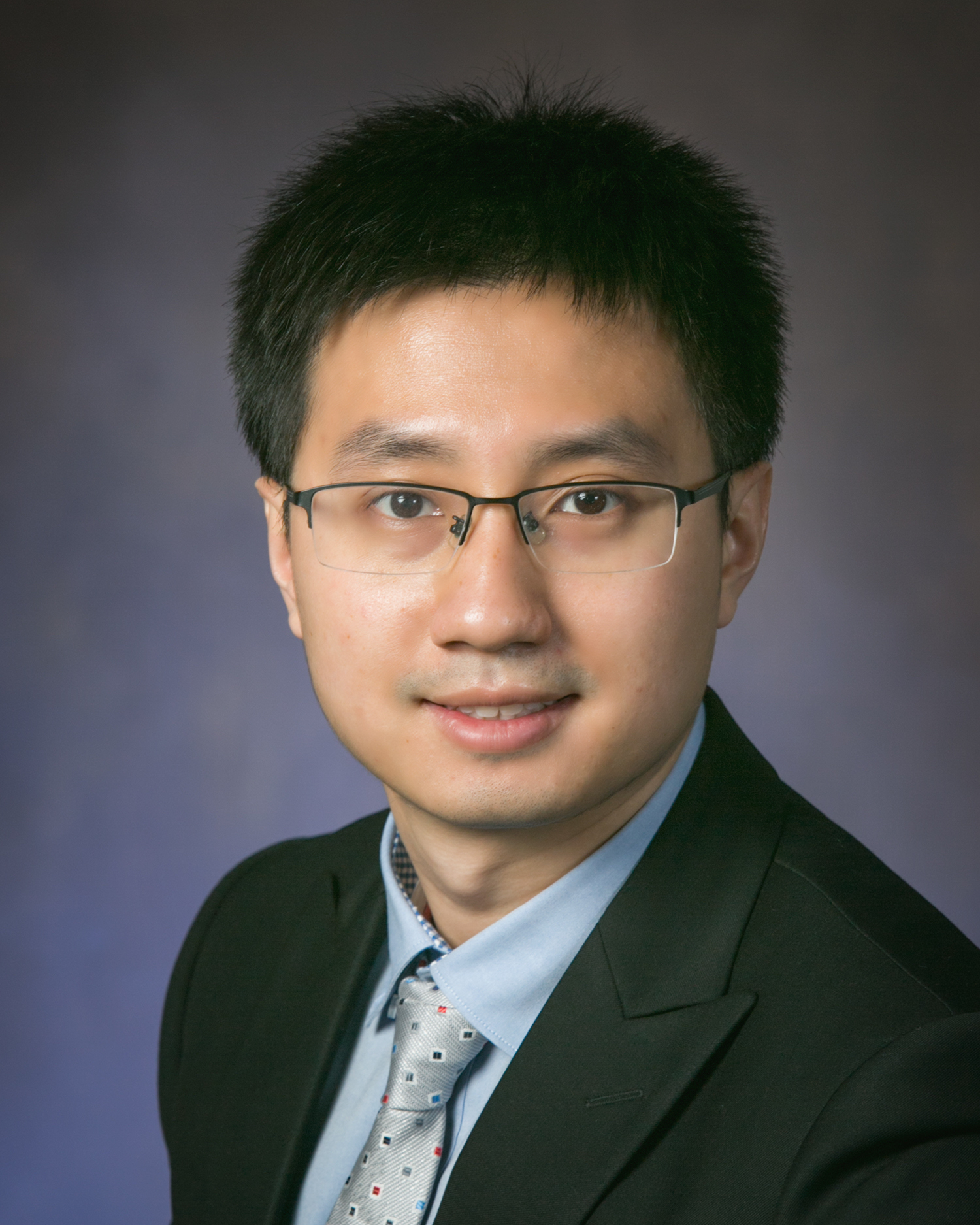}}}]{Yang Zhang}
	received his M.S. and Ph.D. degrees at the University of Illinois at Urbana-Champaign. He is currently a researcher with the MIT-IBM Watson AI Lab. His research interest includes deep learning for speech, audio, and time-series processing.
	\end{IEEEbiography}

	\begin{IEEEbiography}[{\includegraphics[width=1in,height=1.25in,clip,keepaspectratio]{{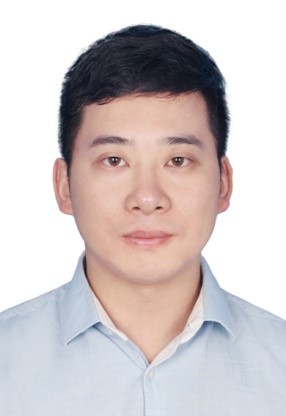}}}]{Mingkui Tan}
	is currently a professor with the School of Software Engineering at South China University of Technology. He received his Bachelor Degree in Environmental Science and Engineering in 2006 and Master degree in Control Science and Engineering in 2009, both from Hunan University in Changsha, China. He received the Ph.D. degree in Computer Science from Nanyang Technological University, Singapore, in 2014. From 2014-2016, he worked as a Senior Research Associate on computer vision in the School of Computer Science, University of Adelaide, Australia. His research interests include machine learning, sparse analysis, deep learning and large-scale optimization.
	\end{IEEEbiography}
	
	\vspace{-5cm}
	\begin{IEEEbiography}[{\includegraphics[width=1in,height=1.25in,clip,keepaspectratio]{{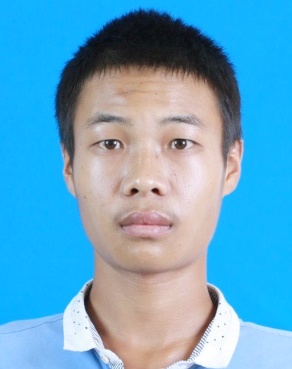}}}]{Hongdong Xiao}
	received the B.E. degree in Computer
    Science and Technology from Nanchang University, China, in 2018.
    He is currently pursuing the M.E. degree at the School of Software Engineering, 
    South China University of Technology, China. His research 
    interests include deep learning in audio understanding.
	\end{IEEEbiography}

	\vspace{-5cm}
	\begin{IEEEbiography}[{\includegraphics[width=1in,height=1.25in,clip,keepaspectratio]{{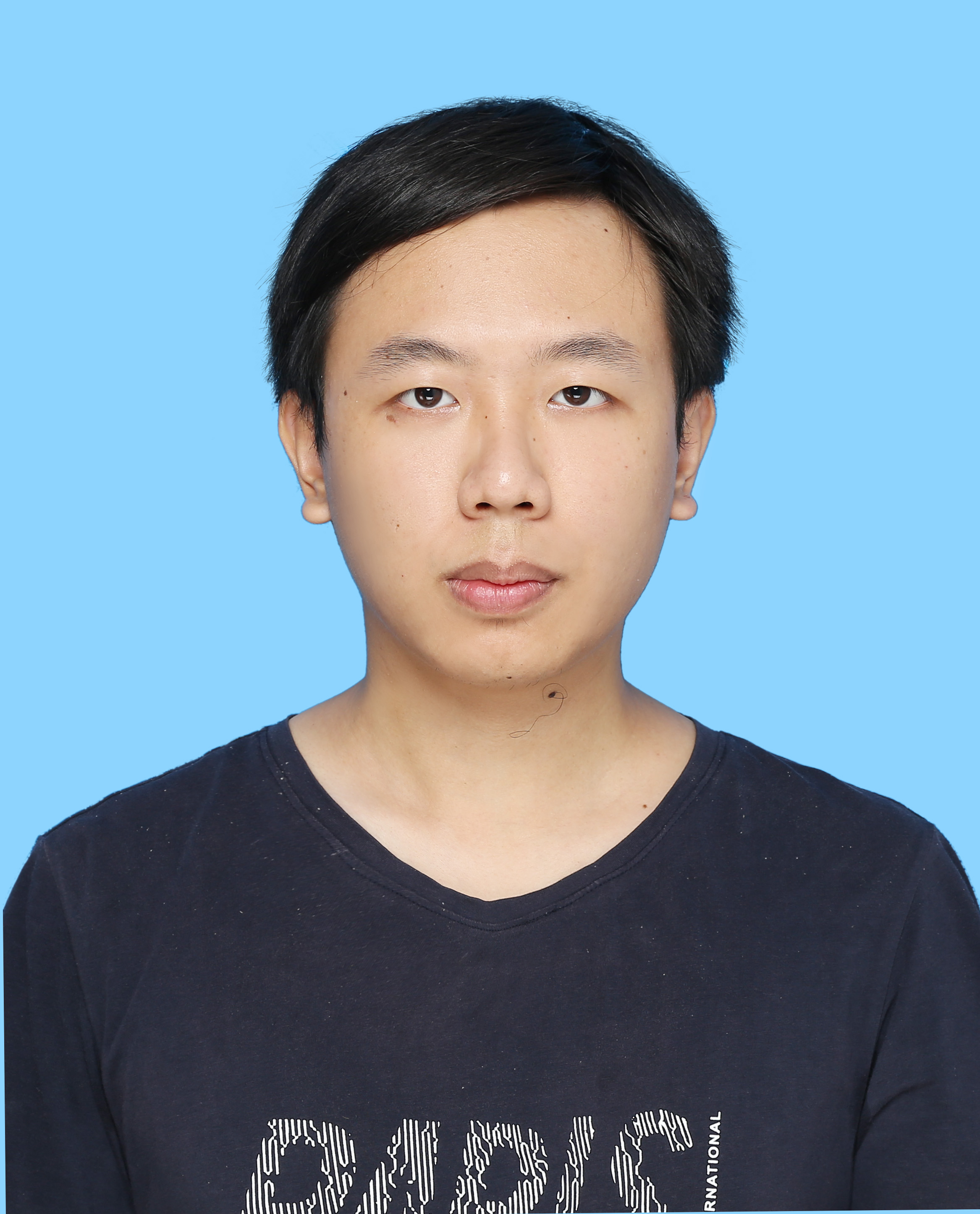}}}]{Deng Huang}
	received the B.E. degree in software engineering from South China University of Technology in 2019. He is currently pursuing the M.E. degree with South China University of Technology. His research interests include computer vision and deep learning.
	\end{IEEEbiography}

	\vspace{-5cm}
	\begin{IEEEbiography}[{\includegraphics[width=1in,height=1.25in,clip,keepaspectratio]{{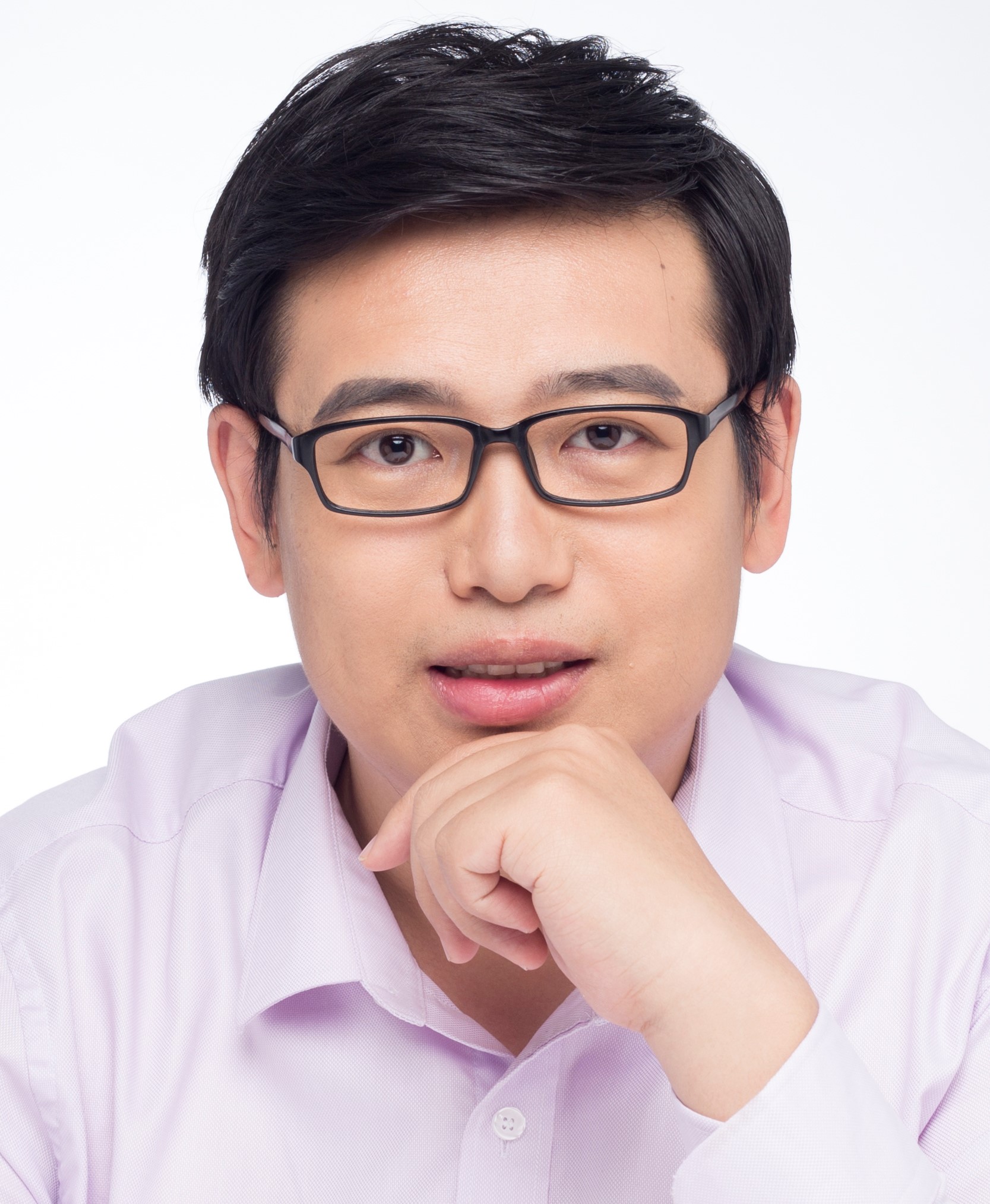}}}]{Chuang Gan}
	is currently a Researcher with the MIT-IBM Watson AI Lab. His research interests mainly include multi-modality learning for video understanding.
	\end{IEEEbiography}

\end{document}